\documentclass[10pt,twocolumn,letterpaper]{article}

\usepackage{cvpr}              
\definecolor{cvprblue}{rgb}{0.21,0.49,0.74}
\usepackage[pagebackref,breaklinks,colorlinks,allcolors=cvprblue]{hyperref}
\usepackage{multirow}
\usepackage{booktabs}
\def\paperID{3051} 
\def\confName{CVPR}
\def\confYear{2026}

\title{
Memory-Augmented Scene Understanding and Exploration for Open-World Aerial Object-Goal Navigation
}

\author{
Jiacong Zhou$^{1}$\footnotemark[1]\quad Jiaxu Miao$^{2}$\footnotemark[1]  \quad Yourun Lin$^{2}$\quad Xianyun Wang$^{2}$ \quad Jun Xiao$^{3}$ \quad Jun Yu$^{2}$\footnotemark[2] \\
$^1$School of Computer Science, Hangzhou Dianzi University\\
$^2$The School of Intelligence Science and Engineering, Harbin
Institute of Technology, Shenzhen \\
$^3$State Key Laboratory of CAD \& CG, Zhejiang University \\
\texttt{jczhou@hdu.edu.cn}
}

\begin{document}
\twocolumn[{
    \renewcommand
    \twocolumn[1][]{#1}
    \maketitle
    \centering
    \vspace{-12pt}
    \includegraphics[width=\textwidth]{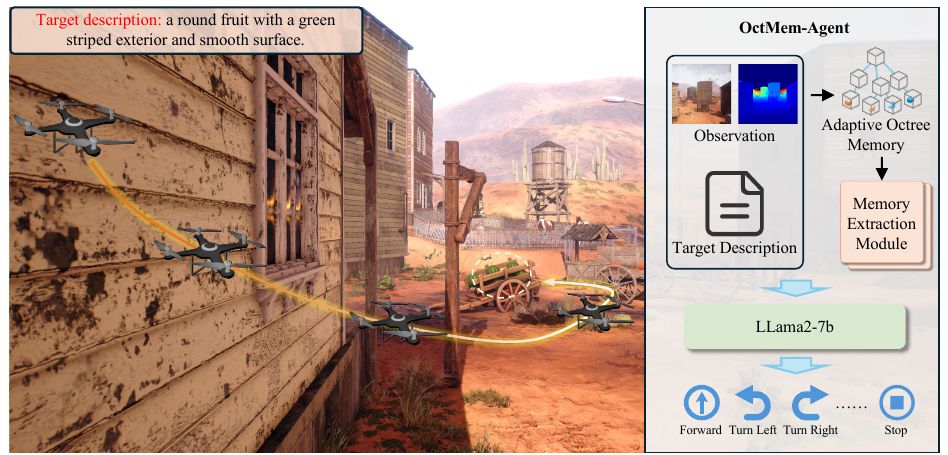}
    \vspace{-1em}
    \captionof{figure}{\small
    Overview of our work. As shown on the left, aerial object-goal navigation requires an Unmanned Aerial Vehicle (UAV) to navigate to target objects in large-scale outdoor environments using only visual observations and high-level object descriptions. In this work, we propose OctMem-Agent, an octree memory-augmented framework illustrated on the right. OctMem-Agent constructs an Adaptive Octree Memory from RGB-D observations and employs a Memory Extraction Module to extract compact information that facilitates navigation decision-making.
}
    \label{fig: teaser}
    \vspace{8pt}
}]

\maketitle

{
  \renewcommand{\thefootnote}%
    {\fnsymbol{footnote}}
  \footnotetext[1]{Equal contribution.}
  \footnotetext[2]{Corresponding author.}
}

\begin{abstract}
Aerial object-goal navigation (Aerial ObjectNav) requires  an Unmanned Aerial Vehicle (UAV) to navigate to target objects in large-scale outdoor environments using only visual observations and high-level object descriptions, without detailed step-by-step instructions. Existing approaches rely on local observations or short-term history, lacking comprehensive scene understanding and effective spatial exploration strategies, which constrains their navigation capability in complex aerial scenarios. To address these challenges, we propose \textbf{OctMem-Agent}, an octree memory-augmented framework for Aerial ObjectNav. Specifically, we introduce an Adaptive Octree Memory Representation that incrementally aggregates RGB-D observations into a hierarchical 3D representation, capturing both explored regions and unexplored frontiers across large-scale aerial environments. We further propose a Instruction-Guided Memory Query module that extracts task-relevant scene and exploration tokens through instruction-modulated queries. By integrating these tokens with visual observations and language instructions, OctMem-Agent achieves comprehensive scene understanding and effective spatial exploration for target localization. Extensive experiments on the Aerial ObjectNav benchmark UAV-ON demonstrate that our method achieves a significant 7.5\% improvement in success rate over existing methods, validating the effectiveness of our design.

\end{abstract}        
\section{Introduction}
\label{sec:intro}

Building on recent advances in aerial Vision-and-Language Navigation (VLN)~\cite{gao2025openfly,gao2024aerial}, Aerial object-goal navigation (Aerial ObjectNav) enables UAVs to navigate and locate target objects based on visual observations and high-level object descriptions~\cite{xiao2025uav}, rather than detailed, step-by-step instructions. This capability is critical for deploying UAVs in emergency response, search and rescue, and package delivery scenarios~\cite{asadzadeh2022uav,zhao2024joint,golabi2021intelligent}. Such applications often feature rapidly changing environmental conditions, unavailable communication infrastructure, and impractical human supervision, requiring UAVs to autonomously locate target objects. However, achieving robust Aerial ObjectNav poses unique challenges. Compared to ground-based ObjectNav~\cite{xia2018gibson,liu2025toponav} in indoor settings, Aerial ObjectNav operates in large-scale outdoor environments characterized by complex terrain and dramatic viewpoint variations~\cite{gao2024aerial}, which demands scale-adaptive scene perception and effective spatial exploration strategies to effectively locate targets.  

Existing methods~\cite{gao2025openfly,xiao2025uav} rely on local observations and short-term history, lacking global scene understanding and long-term spatial memory, thus hindering effective navigation in complex aerial scenarios. For instance, AOA~\cite{xiao2025uav} generates navigation actions via LLMs using current observations and recent pose history, while OpenFly~\cite{gao2025openfly} operates with only the current frame and the previous two frames. This often results in myopic decisions and redundant exploration across large-scale outdoor environments. Furthermore, scene representations developed for ground-based indoor navigation~\cite{krantz2020beyond,xia2018gibson,ramakrishnan2021habitat} (e.g., volumetric maps~\cite{liu2024volumetric}, semantic maps~\cite{liu2025toponav,wang2023gridmm}) cannot be directly applied to aerial scenarios due to drastically larger spatial scales and altitude-dependent viewpoint variations~\cite{zhao2025aerial}.



To address these challenges, we propose OctMem-Agent, an octree memory-augmented vision-language navigation framework for aerial object-goal navigation. Specifically, we introduce an Adaptive Octree Memory Representation that incrementally aggregates RGB-D observations into a hierarchical 3D representation. The memory construction consists of three stages: first projecting 2D visual features into 3D patches with geometric information~\cite{zhu2025llava3d}, then hierarchically aggregating patches using distance-adaptive voxelization that balances fine-grained local details with compact distant representations, and finally integrating the aggregated features into an octree-based memory structure. This hierarchical representation enables effective spatial indexing and scalable memory management across large-scale aerial environments, providing comprehensive scene understanding for long-term navigation.

Building upon this memory representation, we introduce an Instruction-Guided Memory Query module that extracts task-relevant information through instruction-modulated queries. The module initializes two types of learnable queries—scene queries and exploration queries—and employs Feature-wise Linear Modulation (FiLM)~\cite{perez2018film} to condition them on language instructions. These modulated queries then attend to different regions of the memory via a Q-Former~\cite{li2023blip2}: scene queries focus on nearby memory voxels for target localization, while exploration queries attend to distant voxels for identifying unexplored frontiers. By integrating the extracted memory representations with current visual observations and language instructions, OctMem-Agent achieves comprehensive scene understanding and effective spatial exploration. We conduct extensive experiments on the challenging Aerial ObjectNav benchmark UAV-ON~\cite{xiao2025uav}. Results demonstrate that OctMem-Agent achieves a significant 7.5\% improvement in success rate over existing methods, validating the effectiveness of our memory-augmented approach. Overall, the main contributions of our work are as follows:

\begin{itemize}

\item We present an Adaptive Octree Memory that incrementally aggregates RGB-D observations into hierarchical 3D representations for scene understanding in large-scale environments.

\item We introduce an Instruction-Guided Memory Query module that extracts complementary scene and exploration representations through instruction-modulated queries, effectively balancing target localization and frontier exploration.

\item Extensive experiments on Aerial ObjectNav benchmark UAV-ON show that our method achieves state-of-the-art performance with  a significant improvement in success rate.


\end{itemize}

\section{Related Works}

\textbf{\indent Aerial Object-goal Navigation.} Traditional object-goal navigation (ObjectNav)~\cite{anderson2018evaluation} requires an agent to navigate towards a specified object category based on visual observations and language instructions~\cite{mossvln}. Early works primarily focused on indoor environments such as AI2-THOR~\cite{kolve2017ai2}, Gibson~\cite{xia2018gibson}, RoboTHOR~\cite{deitke2020robothor}, and HM3D~\cite{ramakrishnan2021habitat}. Methodologically, these approaches evolved from end-to-end reinforcement learning and imitation learning~\cite{DBLP:conf/iccv/YeBDW21,DBLP:journals/ral/YeY21,DBLP:conf/cvpr/RamrakhyaBWD23} to methods leveraging multimodal large language models~\cite{DBLP:journals/tmlr/ZhangWLLMLL25,liu2023visual,team2023gemini} for semantic understanding and spatial reasoning~\cite{DBLP:conf/rss/ZhangWXZHF0ZW24,wang2025dynam3d,zhang2024uninavid}. More recently, zero-shot ObjectNav methods leverage Vision-Language Models (VLMs) to generalize towards unseen object categories~\cite{zhao2023semantic,DBLP:conf/icra/ZhaoZHQL23,chen2023zero}.

Recently, aerial object-goal navigation (Aerial ObjectNav) has emerged as a new frontier~\cite{xiao2025uav,citynav,zhang2025citynavagent}, presenting unique challenges compared to ground-based navigation. Unlike constrained indoor environments, Aerial ObjectNav operates in large-scale outdoor scenes with complex terrain and dynamic environmental changes, demanding sophisticated spatial reasoning across unrestricted three-dimensional space~\cite{gao2025openfly,WangYWKCW0L025,DBLP:conf/iccv/LiuZQ0ZW23}. Current approaches to aerial object-goal navigation, such as AOA~\cite{xiao2025uav},GeoNav~\cite{xu2025geonav}, primarily focus on zero-shot generalization by leveraging Vision-Language Models (VLMs) to handle diverse high-level goals in unconstrained outdoor environments. In contrast, our method builds upon pretrained VLMs and integrates an Adaptive Octree Memory Representation for scene representation and spatial exploration in complex aerial environments.

\textbf{Memory-based Vision Language Navigation.}
Spatial memory plays a crucial role in vision-language navigation by enabling agents to build and maintain scene representations~\cite{li2024memonav}. Existing spatial memory methods can be broadly categorized into dense representations and sparse representations. Dense representations include early semantic maps~\cite{henry2012rgb,chaplot2020object} built with task-specific networks for limited object categories, as well as more recent voxel maps~\cite{liu2024volumetric} and neural feature fields~\cite{shafiullah2022clip,maggio2024clio} leveraging pretrained vision-language models~\cite{zhang2025novel,huang2022visual,radford2021learningclip}. These methods provide fine-grained spatial encoding but incur substantial memory overhead. Sparse representations, such as topological graphs~\cite{liu2025toponav}, grid maps~\cite{wang2023gridmm,DBLP:conf/acl/ChenLXCLW24}, and hierarchical graph structures~\cite{gu2024conceptgraphs,werby2024hierarchical,wang2025open}, offer more compact scene encoding through structured abstractions.

However, both paradigms face significant challenges when scaling to open-world aerial 3D navigation. Dense representations struggle with large spatial extents, as their memory requirements grow cubically with scene size, making them impractical for outdoor environments with dynamic conditions and diverse terrain. Sparse graph-based representations are limited by their design assumptions. Their discrete spatial abstractions and hierarchical structures, optimized for indoor object-level scene understanding, do not generalize well to large-scale outdoor scenarios.

\begin{figure*}[t]
    \centering
    \includegraphics[width=0.99\textwidth]{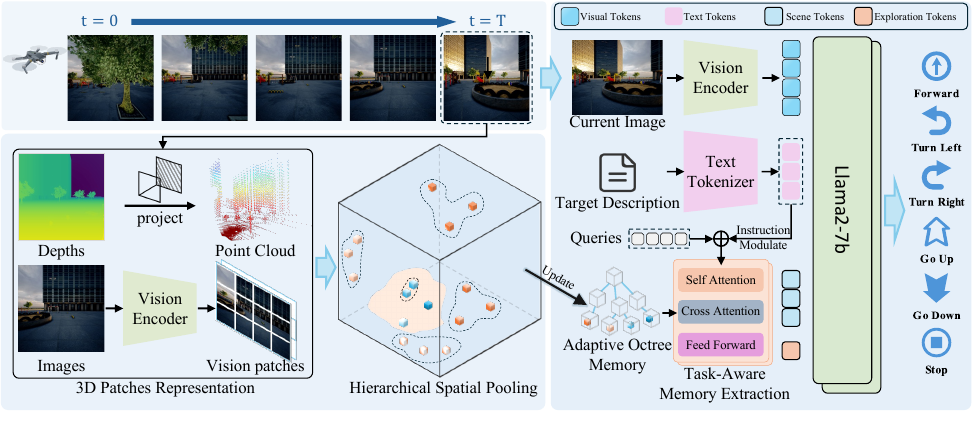}
    \vspace{-1.25em}
    \caption{The overall framework of our proposed OctMem-Agent. The agent builds an Adaptive Octree Memory from historical RGB-D data. The instruction-guided query module extracts compact scene and exploration tokens. These tokens are fused with the current visual observation and language instruction, serving as input to the LLM for action inference.}
    \label{fig:overview}
    \vspace{-1.5em}
\end{figure*}

\section{Method}
\label{method}

In this section, we present \textbf{OctMem-Agent}, a vision-language navigation framework for aerial object-goal navigation. As Figure~\ref{fig:overview} shows, the framework comprises three key components: (1) Adaptive Octree Memory Representation that aggregates historical RGB-D observations into a scalable 3D representation, (2) Instruction-Guided Memory Query that extracts compact scene and exploration tokens via a FiLM-modulated Q-Former, and (3) Action Generation with Memory Integration that combines memory, observation, and language instruction to predict actions.

\subsection{Problem Formulation}
Given a language instruction $I_{\text{goal}} = \{w_1, \dots, w_{N_I}\}$ consisting of $N_I$ words that describes a navigation objective, Aerial ObjectNav aims to guide an aerial agent from an initial 3D pose $p_0 = (x_0, y_0, z_0, \phi_0)$ to the target location by executing a sequence of actions. Here, $(x_0, y_0, z_0) \in \mathbb{R}^3$ denotes the initial position and $\phi_0$ represents the yaw angle. At each time step $t$, the agent receives an observation $O_t = \{\mathbf{D}_t, \mathbf{V}_t\}$, 
where $\mathbf{D}_t$ and $\mathbf{V}_t$ denote the depth map and RGB image respectively, 
and executes an action $a_t \in \mathcal{A}$ conditioned on $I_{\text{goal}}$ and $O_t$ 
to update its pose to $p_{t+1}$. Following~\cite{xiao2025uav,gao2025openfly}, we define the discrete action space as $\mathcal{A} = \{ \allowbreak \texttt{Move~Forward}, \allowbreak \texttt{Turn~Left},\allowbreak \texttt{Turn~Right},\allowbreak \texttt{Ascend},\allowbreak \texttt{Descend}, \texttt{Stop}\}$, where forward movement(\eg,~\texttt{Move~Forward}) translates the agent by 3 units, vertical movement(\eg,~\texttt{Ascend}) adjusts altitude by 3 meters, and turning rotates the agent by 30 degrees. The episode terminates when the agent executes a \texttt{Stop} action or reaches the maximum step limit. Consistent with previous work~\cite{xiao2025uav,gao2025openfly,zhao2025aerial}, the navigation is considered successful if the agent is within 20 meters of the destination.


\subsection{Adaptive Octree Memory Representation}
\label{sec:memory_rep}
To achieve effective long-term spatial representation and 
reasoning in large-scale 3D aerial environments, we propose an \textbf{Adaptive Octree Memory} that aggregates historical observations into a unified and hierarchical memory representation. Our memory construction process consists of three stages: (1) 3D Patches Representation, (2) Hierarchical Spatial Aggregation, and (3) Memory Integration.


\textbf{3D Patches Representation.} To enable better 3D understanding, we construct 3D patches by injecting 3D positional information into 2D patches. This allows the 3D patches to 
explicitly capture spatial structure while retaining the semantic information 
from the original 2D patches. Specifically, for each incoming observation $O_t = \{\mathbf{D}_t, \mathbf{V}_t\}$ at timestep $t$, a vision encoder extracts patch token features from the RGB image $\mathbf{V}_t$ and 
reshapes them into a spatial grid $\mathbf{X}_p \in \mathbb{R}^{h \times w \times d}$. Here $h \times w$ is the spatial resolution and $d$ is the feature dimension. 
Following~\cite{zhu2025llava3d,jain2024odin}, we back-project these 
patches to obtain their 3D world coordinates $\mathbf{P} \in \mathbb{R}^{h \times w \times 3}$ 
using the camera's intrinsic and extrinsic parameters and nearest neighbor 
depth values from $\mathbf{D}_t$. The 3D positions $\mathbf{P}$ are then encoded into 3D positional embeddings $\mathbf{P}' \in \mathbb{R}^{h \times w \times d}$
via a learnable two-layer MLP. We obtain the 3D patches by fusing the positional embeddings with the 2D features:
\begin{equation}
\mathbf{X}_{3D} = \mathbf{X}_p + \mathbf{P}'.
\end{equation}

\textbf{Hierarchical Spatial Aggregation.} The raw 3D patches are often overly dense, making direct processing computationally 
prohibitive. We employ hierarchical pooling to adaptively aggregate spatially important patches.

In Aerial ObjectNav tasks, points beyond the agent's effective planning range are spatially irrelevant as they provide limited navigational value (e.g., points beyond 10000 meters are unlikely to be useful when searching for objects within a 100-meter range). We thus filter out points excessively far from the current agent position $(x_t, y_t, z_t)$. Specifically, we remove points from $\mathbf{P}$ whose distance to $(x_t, y_t, z_t)$ exceeds a threshold $d_{\max}$, and apply the same filtering to $\mathbf{X}_{3D}$ accordingly, obtaining the filtered sets $\hat{\mathbf{P}}$ and $\hat{\mathbf{X}}_{3D}$. Since precise navigation requires detailed local information while effective exploration demands compact representation of distant space, we represent nearby regions with finer-grained voxels to preserve geometric and semantic details, and distant regions with coarser voxels to capture large-scale layout and unexplored frontiers. To achieve this, we partition the filtered points $\hat{\mathbf{P}}$ into hierarchical voxel grids with varying sizes determined by their distance from the agent. Specifically, we define distance-based intervals:
\begin{equation}
\mathcal{D} = \{[d_0, d_1), [d_1, d_2), \ldots, [d_{K-1}, d_K]\},
\label{eq:intervals}
\end{equation}
where $d_0 = 0$ and $d_K = d_{\max}$, and $K$ is the number of intervals. Each interval $[d_{k-1}, d_k)$ corresponds to a voxel size $s_k$ with $s_1 < s_2 < \cdots < s_K$. Within each voxel $v$, both the 3D points and their associated patch features are aggregated via mean pooling:
\begin{equation}
\mathbf{x}_v = \frac{1}{\lvert \mathcal{I}_v \rvert}\sum_{i \in \mathcal{I}_v} \mathbf{x}_i, \quad \mathbf{p}_v = \frac{1}{\lvert \mathcal{I}_v \rvert}\sum_{i \in \mathcal{I}_v} \mathbf{p}_i,
\end{equation}
where $\mathcal{I}_v$ denotes the set of indices in voxel $v$. This yields the final aggregated representations $\mathbf{X}_{\text{agg}} = \{\mathbf{x}_v | v \in \mathcal{V}\}$ and $\mathbf{P}_{\text{agg}} = \{\mathbf{p}_v | v \in \mathcal{V}\}$, where $\mathcal{V}$ is the set of all voxels.

\textbf{Memory Integration.} The aggregated patches $\mathbf{X}_{\text{agg}}$ and points $\mathbf{P}_{\text{agg}}$ are incrementally integrated into a global octree-based memory structure $\mathcal{M}_t$. The octree structure enables effective spatial indexing and hierarchical representation of the 3D environment. Each octree cell corresponds to a spatial region and stores the aggregated patch features of points within that region, allowing fast retrieval and update of local representations.

Initially at $t=0$, we construct the octree memory from the first set of aggregated points $\mathbf{P}_{\text{agg}}^{(0)}$ by recursively partitioning the 3D space into octants based on spatial occupancy. Each point $i$ is assigned to a cell $c$, with the cell feature initialized as $\mathbf{m}_c^{(0)} = \mathbf{x}_i^{(0)}$, forming the initial memory $\mathcal{M}_0 = \{\mathbf{m}_c^{(0)} | c \in \mathcal{C}_0\}$. At subsequent timesteps $t>0$, new points $\mathbf{P}_{\text{agg}}^{(t)}$ and features $\mathbf{X}_{\text{agg}}^{(t)}$ are inserted into the octree by matching them to their corresponding spatial cells based on 3D coordinates. For points matched to existing cells, the corresponding cell features are updated by averaging with their associated features from $\mathbf{X}_{\text{agg}}^{(t)}$. For each point $j$ in previously unvisited regions, a new octree cell $c'$ is created and initialized as $\mathbf{m}_{c'}^{(t)} = \mathbf{x}_j^{(t)}$. This yields the updated memory representation $\mathcal{M}_t = \{\mathbf{m}_c^{(t)} | c \in \mathcal{C}_t\}$, where $\mathcal{C}_t$ denotes the set of all cells (both existing and newly created) at timestep $t$.

\subsection{Instruction-Guided Memory Query}

In this section, we present a method to extract task-relevant information from $\mathcal{M}_t$ based on instruction $I_{\text{goal}}$ via two key components: (1) Instruction-Modulated Query and (2) Task-Aware Memory Extraction.

\textbf{Instruction-Modulated Query.}
We initialize a set of learnable query embeddings $\mathbf{Q} \in \mathbb{R}^{N_q \times d}$, where $N_q$ is the number of queries and $d$ matches the dimension of visual patch embeddings.

To enable task-specific memory querying, we employ Feature-wise Linear Modulation (FiLM)~\cite{perez2018film,kim2025openvlaoft} to inject instruction information into the queries. Specifically, we encode the natural language instruction $I_{\text{goal}}$ into an embedding using a pre-trained language encoder, followed by average pooling to obtain a $d$-dimensional vector $\mathbf{e}_I \in \mathbb{R}^{d}$. Then, we compute feature-wise scaling and shifting parameters $\boldsymbol{\gamma}(\mathbf{e}_I) \in \mathbb{R}^{d}$ and $\boldsymbol{\beta}(\mathbf{e}_I) \in \mathbb{R}^{d}$ via linear projections. 

These parameters modulate each dimension of the initial queries $\mathbf{Q} \in \mathbb{R}^{N_q \times d}$ through an affine transformation:
\begin{equation}
\mathbf{Q}_{\text{task}} = \text{FiLM}(\mathbf{Q}, \mathbf{e}_I) = (1 + \boldsymbol{\gamma}(\mathbf{e}_I)) \odot \mathbf{Q} + \boldsymbol{\beta}(\mathbf{e}_I),
\label{eq:film}
\end{equation}
where $\odot$ denotes element-wise multiplication along each feature dimension. This conditioning ensures that the subsequent memory attention operates with respect to the navigation goal.

\textbf{Task-Aware Memory Extraction.}
The queries $\mathbf{Q}_{\text{task}}$ are then used to extract relevant information from the global memory $\mathcal{M}_t$ via a Q-Former~\cite{li2023blip2}.

To effectively balance target localization and frontier exploration, $\mathbf{Q}_{\text{task}}$ consists of two distinct subsets, each serving complementary purposes:
(1) \textbf{Scene tokens} $\mathbf{Q}_{\text{scene}} \in \mathbb{R}^{N_s \times d}$, which attend to memory voxels within a specified boundary range, focusing on semantic objects that match the target description in $I_{\text{goal}}$ to enable precise target localization.
(2) \textbf{Exploration tokens} $\mathbf{Q}_{\text{explore}} \in \mathbb{R}^{N_e \times d}$, which attend to memory voxels beyond this boundary, identifying frontiers between explored and unexplored regions to guide the agent towards potential search areas when the target is not immediately visible.
Here, $N_s + N_e = N_q$, and both subsets undergo the same FiLM modulation described in Eq.~\ref{eq:film}. We select a distance threshold $d_b$ from the interval set $\mathcal{D}$ (Eq.~\ref{eq:intervals}) to define this boundary range, where scene tokens query voxels within distance $d_b$ from the agent, while exploration tokens query voxels beyond $d_b$.

Within the Q-Former, queries are processed through $L$ Transformer layers, each consisting of self-attention followed by cross-attention with the memory.  During cross-attention, we partition the memory $\mathcal{M}_t$ into two subsets:
\begin{equation}
\begin{aligned}
\mathcal{M}_{\text{near}} &= \{\mathbf{m}^{(t)}_c \in \mathcal{M}_t \mid \text{dis}(c) < d_b\}, \\
\mathcal{M}_{\text{far}} &= \{\mathbf{m}^{(t)}_c \in \mathcal{M}_t \mid \text{dis}(c) \geq d_b\},
\end{aligned}
\label{eq:memory_partition}
\end{equation}
where $\text{dis}(c)$ denotes the Euclidean distance from the agent to cell $c$, and $d_b$ is the boundary distance threshold. Scene tokens $\mathbf{Q}_{\text{scene}}$ perform cross-attention exclusively with $\mathcal{M}_{\text{near}}$ to capture local spatial context, while exploration tokens $\mathbf{Q}_{\text{explore}}$ attend only to $\mathcal{M}_{\text{far}}$ to identify distant unexplored regions. This yields complementary representations $\mathbf{Q}^{(L)}_{\text{scene}}$ and $\mathbf{Q}^{(L)}_{\text{explore}}$ that guide subsequent navigation decisions.

\subsection{Action Generation with Memory Integration}

Our OctMem-Agent builds upon the OpenVLA~\cite{kim2024openvla} framework, adapting it for discrete action prediction in aerial object-goal navigation. To integrate the extracted memory into the model, the Q-Former outputs $\mathbf{Q}^{(L)}_{\text{scene}}$ and $\mathbf{Q}^{(L)}_{\text{explore}}$ are concatenated and projected through a linear layer to match the VLM's hidden dimension. This yields memory representations $\mathbf{H}_{\text{mem}}$. To incorporate the current observation, we encode the RGB image $\mathbf{V}_t$ using the same vision encoder as in Section~\ref{sec:memory_rep} to obtain 2D patch features $\mathbf{H}_{\text{obs}}$. The language instruction $I_{\text{goal}}$ is processed through a pre-trained language encoder to produce text embeddings $\mathbf{H}_{\text{lang}}$. These multimodal representations are concatenated along the sequence dimension to form a unified input sequence $\mathbf{H}_{\text{input}} = [\mathbf{H}_{\text{lang}}, \mathbf{H}_{\text{obs}}, \mathbf{H}_{\text{mem}}]$. Following OpenVLA~\cite{kim2024openvla}, we feed $\mathbf{H}_{\text{input}}$ into a pre-trained vision-language model (VLM) backbone, which processes the sequence through its transformer layers to generate predicted action tokens. Finally, an action decoder discretizes the predicted tokens into discrete actions $a_t \in \mathcal{A}$.
\section{Experiments}
\label{sec:exps}

In this section, we conduct comprehensive experiments on aerial object navigation benchmarks to
evaluate the effectiveness of our OctMem-Agent. Extensive ablation studies are also conducted to show the effectiveness of our Adaptive Octree Memory Representation and the Instruction-Guided Memory Query.

\subsection{Experimental Settings}

\begin{table*}[ht]

    \centering
    \caption{Performance comparison on the UAV-ON benchmark across different target object sizes. Our OctMem-Agent achieves the highest success rates across all object sizes and surpasses all baselines in total performance.}
    \label{tab:uav_on_main_results}
    
    \resizebox{\textwidth}{!}{
        \begin{tabular}{l ccc| ccc| ccc| ccc}
        \hline
        \multirow{2}{*}{\textbf{Method}} & \multicolumn{3}{c}{\textbf{Small}} & \multicolumn{3}{c}{\textbf{Medium}} & \multicolumn{3}{c}{\textbf{Large}} & \multicolumn{3}{c}{\textbf{Total}} \\
        & SR$\uparrow$ & OSR$\uparrow$ & SPL$\uparrow$ & SR$\uparrow$ & OSR$\uparrow$ & SPL$\uparrow$ & SR$\uparrow$ & OSR$\uparrow$ & SPL$\uparrow$ & SR$\uparrow$ & OSR$\uparrow$ & SPL$\uparrow$ \\
        \hline
            Random & 4.14\% & 7.80\% & 2.80\% & 3.33\% & 8.10\% & 3.05\% & 2.48\% & 8.07\% & 1.62\% & 3.70\% & 8.00\% & 2.66\% \\

            CLIP-H~\cite{xiao2025uav} & 2.86\% & 8.43\% & 1.51\% & 10.95\% & 16.67\% & \textbf{7.17}\% & 13.04\% & 19.25\% & 10.53\% & 6.20\% & 11.90\% & 4.15\% \\
            
            AOA-F~\cite{xiao2025uav} & 4.45\% & 16.38\% & 1.61\% & 10.48\% & 17.62\% & \underline{6.36}\% & 14.29\% & 21.74\% & \underline{10.66}\% & 7.30\% & 17.50\% & 4.06\% \\

            AerialVLN~\cite{gao2024aerial} & 4.13\% & 19.87\% & 4.01\% & 4.76\% & 17.62\% & 4.29\% & 1.86\% & 19.25\% & 1.86\% & 3.90\% & 19.30\% & 3.72\% \\

            Navid~\cite{zhang2024navid} & 10.02\% & \underline{29.55}\% & 4.41\% & \underline{11.43}\% & \underline{26.44}\% & 5.95\% & \underline{17.39}\% & 30.73\% & \textbf{15.03}\% & 11.50\% & \underline{29.10}\% & \textbf{6.44}\% \\

            OpenFly~\cite{gao2025openfly} & \underline{12.40}\% & 26.23\% & \textbf{6.35}\% & 10.00\% & 20.48\% & 4.19\% & 13.04\% & \underline{31.65}\% & 7.56\% & \underline{12.00}\% & 25.90\% & {6.09}\% \\

         \hline
            OctMem-Agent & \textbf{18.91}\% & \textbf{29.57}\% & \underline{6.03}\% & \textbf{18.09}\% & \textbf{26.66}\% & 6.27\% & \textbf{23.60}\% & \textbf{31.67}\% & 7.80\% & \textbf{19.50}\% & \textbf{29.30}\% & \underline{6.37}\% \\
        \hline
        \end{tabular}
}
\vspace{-1.5em}
\end{table*}

\textbf{\indent Benchmark.} 
We primarily evaluate our method on UAV-ON~\cite{xiao2025uav}, a comprehensive benchmark designed for aerial object-goal navigation in complex and diverse environments. The benchmark contains 14 high-fidelity outdoor environments developed in Unreal Engine, spanning diverse geographic structures from compact urban parks to expansive mountainous terrains, including villages, towns, cities, forests, and water areas. It contains 1,270 target objects with corresponding high-level instructions, with the objects varying in size from small to large.

The benchmark comprises 10,000 training episodes across 10 environments and 1,000 test episodes spanning the 10 training environments and 4 additional held-out environments. The test split includes novel objects and diverse instructions to assess generalization capabilities. Each episode consists of a random starting position and a goal object described by a natural language instruction. Since UAV-ON provides only annotated data without trajectory information, we generate training trajectories following the approaches described in~\cite{xiao2025uav}, which employs a three-dimensional A* algorithm~\cite{hart1968formal} to compute the shortest traversable path to each target object. We then use these trajectories to train both OctMem-Agent and baseline methods.


\textbf{Baselines.} We compare our approach against two categories of methods. (i) We adapt existing aerial vision-language navigation models to the Aerial ObjectNav task on UAV-ON, including aerial navigation methods (OpenFly~\cite{gao2025openfly}, AerialVLN~\cite{gao2024aerial}) and a state-of-the-art indoor VLN model (Navid~\cite{zhang2024navid}). These methods are trained on the UAV-ON training set. (ii) We compare against zero-shot aerial object-goal navigation approaches, including CLIP-H~\cite{xiao2025uav} and AOA-F~\cite{xiao2025uav}, which leverage vision-language foundation models for decision-making. Additionally, we include a random baseline that randomly selects actions at each step until the $\texttt{Stop}$ action is chosen or the maximum step limit is reached, serving as a reference. We provide additional experimental results in the Appendix.

\begin{table}[t]
\centering
\caption{Performance comparison on seen and unseen environments. Our method consistently outperforms baselines on SR and OSR across both settings.}
\label{tab:seen_unseen}
    \resizebox{\linewidth}{!}{
        \begin{tabular}{lcccccc}
            \toprule
            \multirow{2}{*}{\textbf{Method}} & \multicolumn{3}{c}{\textbf{Seen}} & \multicolumn{3}{c}{\textbf{Unseen}} \\
            \cmidrule(lr){2-4} \cmidrule(lr){5-7}
            & \textbf{SR}$\uparrow$ & \textbf{OSR}$\uparrow$ & \textbf{SPL}$\uparrow$ & \textbf{SR}$\uparrow$ & \textbf{OSR}$\uparrow$ & \textbf{SPL}$\uparrow$ \\
            \midrule
            AerialVLN   & 4.55\%        &16.84\%        &4.24\%     &3.51\%   &20.77\%  &3.41\% \\
            Navid       & 14.17\%       & 30.08\%       &7.19\%     &10.70\%  &28.51\%  & \textbf{5.99}\% \\
    
            OpenFly     & 12.29\%           & 26.20\% & 6.83\%    & 11.82\%           & 25.71\%           & 5.64\% \\
            OctMem-Agent& \textbf{22.72}\%  &\textbf{30.48}\%   & \textbf{8.40}\%   & \textbf{17.57}\%  & \textbf{28.59}\% & 5.15\% \\
            \bottomrule
        \end{tabular}
    }
\vspace{-1.75em}
\end{table}

\textbf{Metrics.} We evaluate agent performance using three metrics: Success Rate (SR), Oracle Success Rate (OSR), and Success-weighted Path Length (SPL). These metrics are widely used in embodied navigation tasks~\cite{ramakrishnan2021habitat,zhang2024uninavid}. The detailed definitions are as follows:
\begin{itemize}
\item \textbf{Success Rate (SR)} measures whether the agent stops within a threshold distance from the goal object. Following~\cite{xiao2025uav,citynav}, an episode is considered successful if the final distance between the agent and the target is less than the success threshold of $\tau = 20$ meters.

\item \textbf{Oracle Success Rate (OSR)} measures whether the agent has ever come within the success threshold during the episode, regardless of when it executes the Stop action. This metric captures the agent's capacity to explore and reach near the goal during its trajectory.

\item \textbf{Success-weighted Path Length (SPL)} is a weighted success rate where each successful episode is weighted by the ratio of the shortest path length to the actual trajectory length, measuring both navigation success and path efficiency.

\end{itemize}


\textbf{Implementation details.} We build our method upon OpenVLA~\cite{kim2024openvla} and initialize it from the same pretrained weights as OpenFly~\cite{gao2025openfly} for fair comparison. Consistent with OpenVLA, our shared encoder uses a vision encoder that concatenates features from DINOv2~\cite{DBLP:journals/tmlr/dinov2} and SigLIP~\cite{DBLP:siglip}, while adopting LLaMA-2 7B~\cite{touvron2023llama2openfoundation} as the LLM backbone. Statistical analysis of the UAV-ON benchmark reveals that distances between target objects and initial positions are consistently below 500 meters. Based on this observation, we set $d_{\max}=500$. We set the frontier partitioning distance threshold to $d_b=50$ and establish distance intervals $\mathcal{D}={[0,d_b),[d_b,d_{\text{max}})}$ with corresponding step sizes $s_k \in \{5, 25\}$. For the query settings, we use a total of $N_q=144$ task queries, consisting of $N_s=128$ scene queries and $N_e=16$ explore queries.  We train OctMem-Agent and baselines with a batch size of 64 and a learning rate of 2e-5 for 2 epochs on four L20 GPUs.

\begin{table}[t]
    \centering
    \caption{Ablation study on key components. We evaluate the impact of Adaptive Octree Memory Representation and Instruction-Guided Memory Query on model performance.}
    \label{tab:components}
    \resizebox{0.99\linewidth}{!}{
        \begin{tabular}{cccccccc}
        \toprule
    
        \parbox[c][3em][c]{2.5cm}{\centering\textbf{Adaptive Octree Memory Representation}} & 
        \parbox[c][3em][c]{2.5cm}{\centering\textbf{Instruction-Guided Memory Query}} & 
        \parbox[c][3em][c]{1cm}{\centering\textbf{SR}$\uparrow$} & 
        \parbox[c][3em][c]{1cm}{\centering\textbf{OSR}$\uparrow$} & 
        \parbox[c][3em][c]{1cm}{\centering\textbf{SPL}$\uparrow$} \\
        \midrule
        $\times$    & $\times$      & 12.40\%       & 20.60\%      & 3.35\%  \\
        \checkmark  & $\times$      & 15.70\%       & 21.10\%      & 4.35\% \\
        \checkmark  & \checkmark    & \textbf{19.50}\%       & \textbf{29.30}\%      & \textbf{6.37}\% \\
        \bottomrule
        \end{tabular}
    }
    \vspace{-1.75em}
\end{table}

\begin{figure*}[ht]
    \centering
    \includegraphics[width=0.99\textwidth]{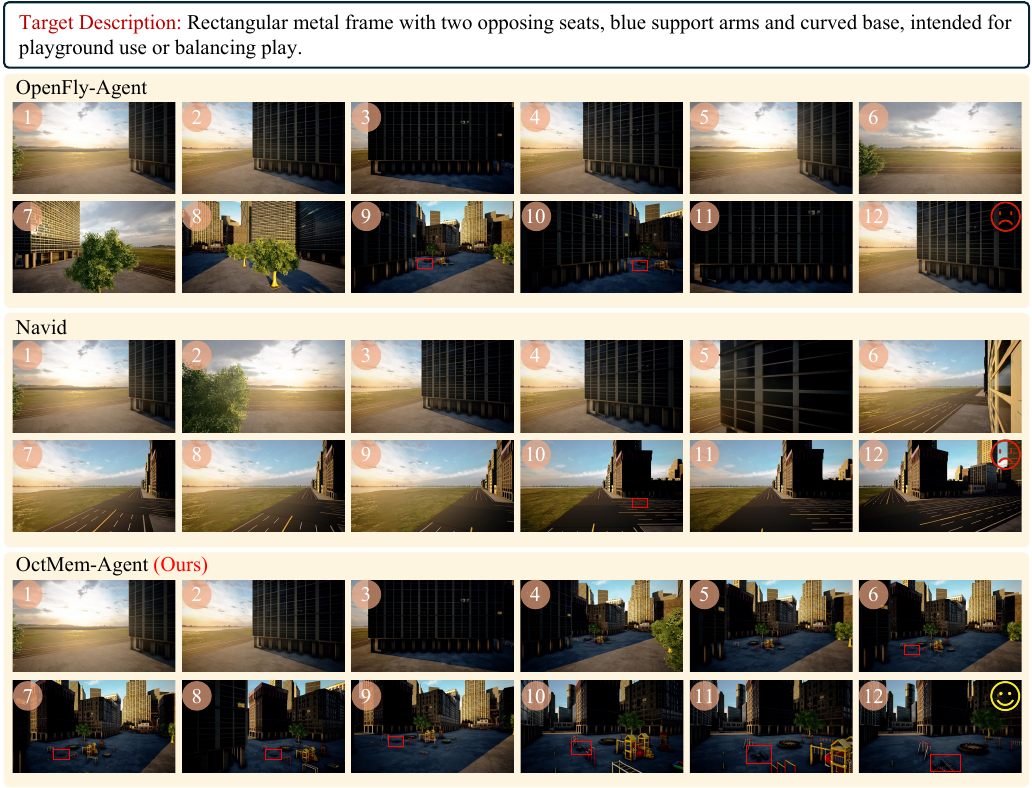} 
    \vspace{-0.25em}
    \caption{We present a qualitative comparison of our OctMem-Agent and baseline methods on UAV-ON. OctMem-Agent maintains consistent detection throughout the entire sequence and successfully reaches the goal, while OpenFly-Agent gets stuck rotating in place and loses the target, and Navid fails to locate the object entirely. This highlights the effectiveness of our OctMem-Agent in achieving robust tracking and successful navigation.}
    \label{fig:case}
    \vspace{-1.75em}
\end{figure*}

\subsection{Main Results}
\textbf{Performance Comparison.} Table \ref{tab:uav_on_main_results} presents a comprehensive performance comparison of our OctMem-Agent against existing methods across different target object sizes (Small, Medium, and Large). Our OctMem-Agent demonstrates superior performance across all baseline categories and object sizes. Compared to zero-shot object-goal navigation approaches, our method significantly outperforms CLIP-H by 13.30\% in SR and 17.40\% in OSR, and surpasses AOA-F by 12.20\% in SR and 11.80\% in OSR on the total results. When compared to trained aerial vision-language navigation models, OctMem-Agent achieves 7.50\% and 8.00\% higher SR than state-of-the-art methods OpenFly~\cite{gao2025openfly} and Navid~\cite{zhang2024navid}, with corresponding improvements of 3.40\% and 0.20\% in OSR, respectively. Notably, our method achieves the highest success rates across all object sizes (Small, Medium, and Large) and surpasses all baselines in total performance, validating the effectiveness of our framework. Our method also achieves favorable results in the SPL metric. These results validate the effectiveness of our approach in handling complex aerial object navigation tasks across diverse target object sizes. 

\textbf{Generalization Capability Analysis.}  We validate our method's generalization capability on the UAV-ON benchmark, which evaluates performance on both seen environments and unseen environments that contain novel objects and diverse instructions. As shown in Table~\ref{tab:seen_unseen}, OctMem-Agent demonstrates robust generalization across multiple metrics. On seen environments, our method achieves superior performance with 22.72\% SR, 30.48\% OSR, and 8.4\% SPL, significantly outperforming all baseline methods. More importantly, on unseen environments, OctMem-Agent maintains its advantage with 17.57\% SR and 28.59\% OSR, consistently surpassing other approaches. This consistent superiority in both seen and unseen settings demonstrates that our OctMem-Agent effectively captures transferable spatial knowledge and navigation strategies. The strong performance on unseen environments validates our method's ability to generalize to novel scenarios with previously unseen target objects and diverse environmental configurations.

\textbf{Case Study.} We present a qualitative comparison of our OctMem-Agent and baseline methods on the UAV-ON benchmark. As shown in Figure~\ref{fig:case}, our OctMem-Agent first explores the surrounding environment, then gradually narrows down potential recreational areas before precisely locating the target and successfully reaching the goal (frame 12). In contrast, OpenFly-Agent gets stuck rotating in place (frames 9-10), losing the target due to poor localization capability. Navid fails to locate the object and moves forward into an open field, revealing insufficient exploration capability. These results demonstrate our method's ability to simultaneously maintain spatial memory for target localization and exploration in large-scale complex environments. 

\subsection{Ablation Studies}

\begin{table}[ht]
    \centering
    \caption{Ablation study on spatial aggregation methods. We compare our proposed Hierarchical Spatial Aggregation with standard voxelization baseline.}
    \label{tab:spatial_aggregation}
    \resizebox{0.98\linewidth}{!}{
        \begin{tabular}{lccc}
            \toprule
            Spatial Aggregation Method & SR & OSR & SPL \\
            \midrule
            Voxelization & 19.10\% & 27.50\% & 5.70\% \\
            Hierarchical Spatial Aggregation & \textbf{19.50\%} & \textbf{29.30\%} & \textbf{6.37\%} \\ 
            \bottomrule
        \end{tabular}
    }
    \vspace{-1.25em}
\end{table}

\subsubsection{Ablation Study on Main Components}

To validate the effectiveness of our proposed components, we conduct an ablation study as shown in Table~\ref{tab:components}. The baseline model achieves 12.40\% SR, 20.60\% OSR, and 3.35\% SPL. Introducing the Adaptive Octree Memory Representation yields consistent improvements across all metrics (15.70\% SR, 21.10\% OSR, 4.35\% SPL), demonstrating that multi-scale spatial encoding captures both fine-grained details and global scene structure essential for navigation. Adding Instruction-Guided Memory Query on top of the memory representation brings additional gains, achieving 19.50\% SR (+3.80\%), 29.30\% OSR (+8.20\%), and 6.37\% SPL (+2.02\%). This indicates that instruction-guided queries effectively extract task-relevant information from the adaptive octree memory representation, reducing unnecessary exploration. The improvements validate the effectiveness of each component in our framework.

\subsubsection{Ablation Study on Adaptive Octree Memory Representation}

We compare our Hierarchical Spatial Aggregation with standard voxelization in Table~\ref{tab:spatial_aggregation}. Standard voxelization partitions 3D space into uniform grid cells for feature aggregation. We set its voxel size to 5 to match the finest granularity in our hierarchical method. While voxelization achieves competitive performance (19.10\% SR, 27.50\% OSR), our hierarchical approach demonstrates notable improvements, particularly in OSR (29.30\%, +1.80\%) and SPL (6.37\%, +0.67\%). By maintaining fine-grained details for nearby regions while abstracting distant areas, the hierarchical representation enables the agent to make more informed local decisions while maintaining global scene awareness. The gains in OSR and SPL indicate that hierarchical aggregation improves the quality of navigation trajectories.

\begin{table}[ht!]
    \centering
    \caption{Ablation study on Instruction-Guided Memory Query. We compare performance with and without Instruction-Modulated Query (IMQ).}
    
    \label{tab:memory-extraction}
    \resizebox{0.95\linewidth}{!}{
        \begin{tabular}{lccc}
            \toprule
            Method & SR (\%) & OSR (\%) & SPL (\%) \\
            \midrule
            OctMem-Agent$_{\text{w/o IMQ}}$ & 18.60 & 27.30 & 5.97 \\
            OctMem-Agent & \textbf{19.50} & \textbf{29.30} & \textbf{6.37}\\
            \bottomrule
        \end{tabular}
    }
    \vspace{-1.25em}
\end{table}

\subsubsection{Ablation Study on Instruction-Guided Memory Query}

To validate the effectiveness of instruction modulation in our memory query mechanism, we conduct an ablation study by comparing the full model with a variant that uses vanilla learnable queries without FiLM-based instruction conditioning. As shown in Table~\ref{tab:memory-extraction}, removing the Instruction-Modulated Query (IMQ) leads to a performance drop across all metrics. Specifically, OctMem-Agent$_{\text{w/o IMQ}}$ achieves 18.60\% SR, 27.30\% OSR, and 5.97\% SPL, while the full OctMem-Agent achieves 19.50\% SR, 29.30\% OSR, and 6.37\% SPL. The improvements of 0.9\%, 2.0\%, and 0.4\% in SR, OSR, and SPL respectively demonstrate that instruction modulation through FiLM is crucial for conditioning queries on navigation goals. This validates that affine transformation of queries based on instruction embeddings enables more effective extraction of task-relevant information from the memory.
\section{Limitations and Conclusion}

In this work, we propose OctMem-Agent, a memory-augmented vision-language navigation framework for aerial object-goal navigation. First, we introduce an Adaptive Octree Memory Representation that incrementally aggregates RGB-D observations into hierarchical 3D representations, enabling effective spatial indexing and scalable memory management across large-scale aerial environments. Second, we develop an Instruction-Guided Memory Query module that extracts task-relevant information through instruction-modulated queries, effectively balancing target localization and frontier exploration. Extensive experiments on the UAV-ON benchmark demonstrate significant improvements over existing methods, validating the effectiveness of our memory-augmented approach. However, our adaptive octree memory representation relies on accurate depth estimation or pose and IMU drift, which may degrade in challenging outdoor conditions such as low-light environments, adverse weather, or textureless regions, leading to erroneous depth estimates. Future work will explore robust depth fusion techniques and multi-sensor integration to improve reliability in diverse environmental conditions.

\section{Acknowledgments}
This work was supported by the National Natural Science Foundation of China (NSFC) under Grant No. 62125201, U24B20174, No. 62572147.

{
    \small
    \bibliographystyle{ieeenat_fullname}
    \bibliography{main}

@String(CVPR= {IEEE Conf. Comput. Vis. Pattern Recog.})

@String(ICCV= {Int. Conf. Comput. Vis.})

@String(ECCV= {Eur. Conf. Comput. Vis.})

@String(ICLR = {Int. Conf. Learn. Represent.})

@String(AAAI = {AAAI})

@String(CVPR  = {CVPR})

@String(ICCV  = {ICCV})

@String(ECCV  = {ECCV})

@String(ICLR  = {ICLR})

@ARTICLE{mossvln,
  author={Yu, Ting and Wu, Yifei and Cui, Qiongjie and Huang, Qingming and Yu, Jun},
  journal={IEEE Transactions on Multimedia}, 
  title={MossVLN: Memory-Observation Synergistic System for Continuous Vision-Language Navigation}, 
  year={2025},
}

@article{anderson2018evaluation,
  title={On evaluation of embodied navigation agents},
  author={Anderson, Peter and Chang, Angel and Chaplot, Devendra Singh and Dosovitskiy, Alexey and Gupta, Saurabh and Koltun, Vladlen and Kosecka, Jana and Malik, Jitendra and Mottaghi, Roozbeh and Savva, Manolis and others},
  journal={arXiv preprint arXiv:1807.06757},
  year={2018}
}

@article{kolve2017ai2,
  title={Ai2-thor: An interactive 3d environment for visual ai},
  author={Kolve, Eric and Mottaghi, Roozbeh and Han, Winson and VanderBilt, Eli and Weihs, Luca and Herrasti, Alvaro and Deitke, Matt and Ehsani, Kiana and Gordon, Daniel and Zhu, Yuke and others},
  journal={arXiv preprint arXiv:1712.05474},
  year={2017}
}

@inproceedings{xia2018gibson,
  title={Gibson env: Real-world perception for embodied agents},
  author={Xia, Fei and Zamir, Amir R and He, Zhiyang and Sax, Alexander and Malik, Jitendra and Savarese, Silvio},
  booktitle={CVPR},
  year={2018}
}

@article{ramakrishnan2021habitat,
  title={Habitat-matterport 3d dataset (hm3d): 1000 large-scale 3d environments for embodied ai},
  author={Ramakrishnan, Santhosh K and Gokaslan, Aaron and others},
  journal={arXiv preprint arXiv:2109.08238},
  year={2021}
}

@inproceedings{deitke2020robothor,
  title={Robothor: An open simulation-to-real embodied ai platform},
  author={Deitke, Matt and Han, Winson and Herrasti, Alvaro and others},
  booktitle={CVPR},
  year={2020}
}

@article{xiao2025uav,
  title={Uav-on: A benchmark for open-world object goal navigation with aerial agents},
  author={Xiao, Jianqiang and Sun, Yuexuan and Shao, Yixin and Gan, Boxi and Liu, Rongqiang and Wu, Yanjing and Guan, Weili and Deng, Xiang},
  journal={arXiv preprint arXiv:2508.00288},
  year={2025}
}

@article{henry2012rgb,
  title={RGB-D mapping: Using Kinect-style depth cameras for dense 3D modeling of indoor environments},
  author={Henry, Peter and Krainin, Michael and Herbst, Evan and Ren, Xiaofeng and Fox, Dieter},
  journal={The international journal of Robotics Research},
  year={2012},
}

@article{chaplot2020object,
  title={Object goal navigation using goal-oriented semantic exploration},
  author={Chaplot, Devendra Singh and Gandhi, Dhiraj Prakashchand and Gupta, Abhinav and Salakhutdinov, Russ R},
  journal={Neurips},
  year={2020}
}

@inproceedings{gu2024conceptgraphs,
  title={Conceptgraphs: Open-vocabulary 3d scene graphs for perception and planning},
  author={Gu, Qiao and Kuwajerwala, Ali and Morin, Sacha and Jatavallabhula, Krishna Murthy and Sen, Bipasha and Agarwal, Aditya and Rivera, Corban and Paul, William and Ellis, Kirsty and Chellappa, Rama and others},
  booktitle={ICRA},
  year={2024},
}

@article{maggio2024clio,
  title={Clio: Real-time task-driven open-set 3d scene graphs},
  author={Maggio, Dominic and Chang, Yun and Hughes, Nathan and Trang, Matthew and Griffith, Dan and Dougherty, Carlyn and Cristofalo, Eric and Schmid, Lukas and Carlone, Luca},
  journal={IEEE Robotics and Automation Letters},
  year={2024},
}

@article{huang2022visual,
  title={Visual language maps for robot navigation},
  author={Huang, Chenguang and Mees, Oier and Zeng, Andy and Burgard, Wolfram},
  journal={arXiv preprint arXiv:2210.05714},
  year={2022}
}

@article{shafiullah2022clip,
  title={Clip-fields: Weakly supervised semantic fields for robotic memory},
  author={Shafiullah, Nur Muhammad Mahi and Paxton, Chris and Pinto, Lerrel and Chintala, Soumith and Szlam, Arthur},
  journal={arXiv preprint arXiv:2210.05663},
  year={2022}
}

@article{gao2025openfly,
  title={OpenFly: A versatile toolchain and large-scale benchmark for aerial vision-language navigation},
  author={Gao, Yunpeng and Li, Chenhui and You, Zhongrui and Liu, Junli and Li, Zhen and Chen, Pengan and Chen, Qizhi and Tang, Zhonghan and Wang, Liansheng and Yang, Penghui and others},
  journal={arXiv e-prints},
  year={2025}
}

@inproceedings{WangYWKCW0L025,
  author       = {Xiangyu Wang and
                  Donglin Yang and
                  Ziqin Wang and others},
  title        = {Towards Realistic {UAV} Vision-Language Navigation: Platform, Benchmark,
                  and Methodology},
  booktitle    = {ICLR},
  year         = {2025},
}

@inproceedings{DBLP:conf/iccv/LiuZQ0ZW23,
  author       = {Shubo Liu and
                  Hongsheng Zhang and
                  Yuankai Qi and
                  Peng Wang and
                  Yanning Zhang and
                  Qi Wu},
  title        = {AerialVLN: Vision-and-Language Navigation for UAVs},
  booktitle    = {ICCV},
  year         = {2023},
}

@inproceedings{citynav,
  author       = {Jungdae Lee and
                  Taiki Miyanishi and
                  Shuhei Kurita and
                  Koya Sakamoto and
                  Daichi Azuma and
                  Yutaka Matsuo and
                  Nakamasa Inoue},
  title        = {CityNav: Language-Goal Aerial Navigation Dataset with Geographic Information},
  booktitle    = {ICCV},
  year         = {2025},
}

@inproceedings{DBLP:conf/cvpr/RamrakhyaBWD23,
  author       = {Ram Ramrakhya and
                  Dhruv Batra and
                  Erik Wijmans and
                  Abhishek Das},
  title        = {PIRLNav: Pretraining with Imitation and {RL} Finetuning for {OBJECTNAV}},
  booktitle    = {CVPR},
  year         = {2023},
}

@inproceedings{DBLP:conf/iccv/YeBDW21,
  author       = {Joel Ye and
                  Dhruv Batra and
                  Abhishek Das and
                  Erik Wijmans},
  title        = {Auxiliary Tasks and Exploration Enable ObjectGoal Navigation},
  booktitle    = {ICCV},
  year         = {2021},
}

@article{DBLP:journals/ral/YeY21,
  author       = {Xin Ye and
                  Yezhou Yang},
  title        = {Efficient Robotic Object Search Via {HIEM:} Hierarchical Policy Learning
                  With Intrinsic-Extrinsic Modeling},
  journal      = {{IEEE} Robotics Autom. Lett.},
  year         = {2021},
}

@article{chen2023zero,
  title={Zero-shot object searching using large-scale object relationship prior},
  author={Chen, Hongyi and Xu, Ruinian and Cheng, Shuo and Vela, Patricio A and Xu, Danfei},
  journal={arXiv preprint arXiv:2303.06228},
  year={2023}
}

@inproceedings{DBLP:conf/icra/ZhaoZHQL23,
  author       = {Qianfan Zhao and
                  Lu Zhang and
                  Bin He and
                  Hong Qiao and
                  Zhiyong Liu},
  title        = {Zero-Shot Object Goal Visual Navigation},
  booktitle    = {ICRA},
  year         = {2023},
}

@article{zhao2023semantic,
  title={Semantic policy network for zero-shot object goal visual navigation},
  author={Zhao, Qianfan and Zhang, Lu and He, Bin and Liu, Zhiyong},
  journal={IEEE Robotics and Automation Letters},
  year={2023},
}

@inproceedings{DBLP:conf/rss/ZhangWXZHF0ZW24,
  author       = {Jiazhao Zhang and
                  Kunyu Wang and
                  Rongtao Xu and
                  Gengze Zhou and
                  Yicong Hong and
                  Xiaomeng Fang and
                  Qi Wu and
                  Zhizheng Zhang and
                  He Wang},
  title        = {NaVid: Video-based {VLM} Plans the Next Step for Vision-and-Language
                  Navigation},
  booktitle    = {RSS},
  year         = {2024},
}

@inproceedings{wang2025dynam3d,
  title={Dynam3D: Dynamic Layered 3D Tokens Empower VLM for Vision-and-Language Navigation},
  author={Wang, Zihan and Lee, Seungjun and Lee, Gim Hee},
  booktitle    = {Neurips},
  year         = {2025},
}

@article{zhang2024uninavid,
    title={Uni-NaVid: A Video-based Vision-Language-Action Model for Unifying Embodied Navigation Tasks}, 
    author={Jiazhao Zhang and Kunyu Wang and Shaoan Wang and Minghan Li and Haoran Liu and Songlin Wei and Zhongyuan Wang and Zhizheng Zhang and He Wang},
    year={2024},
    journal = {arXiv preprint arXiv:2412.06224}
}

@article{DBLP:journals/tmlr/ZhangWLLMLL25,
  author       = {Yuanhan Zhang and
                  Jinming Wu and
                  Wei Li and
                  Bo Li and
                  Zejun Ma and
                  Ziwei Liu and
                  Chunyuan Li},
  title        = {LLaVA-Video: Video Instruction Tuning With Synthetic Data},
  journal      = {Trans. Mach. Learn. Res.},
  year         = {2025},
}

@article{liu2023visual,
  title={Visual instruction tuning},
  author={Liu, Haotian and Li, Chunyuan and Wu, Qingyang and Lee, Yong Jae},
  journal={Advances in neural information processing systems},
  year={2023}
}

@article{team2023gemini,
  title={Gemini: a family of highly capable multimodal models},
  author={Team, Gemini and Anil, Rohan and Borgeaud, Sebastian and Alayrac, Jean-Baptiste and Yu, Jiahui and Soricut, Radu and Schalkwyk, Johan and Dai, Andrew M and Hauth, Anja and Millican, Katie and others},
  journal={arXiv preprint arXiv:2312.11805},
  year={2023}
}

@article{zhao2025aerial,
  title={Aerial Vision-and-Language Navigation with Grid-based View Selection and Map Construction},
  author={Zhao, Ganlong and Li, Guanbin and Pan, Jia and Yu, Yizhou},
  journal={arXiv preprint arXiv:2503.11091},
  year={2025}
}

@article{gao2024aerial,
  title={Aerial vision-and-language navigation via semantic-topo-metric representation guided LLM reasoning},
  author={Gao, Yunpeng and Wang, Zhigang and Jing, Linglin and Wang, Dong and Li, Xuelong and Zhao, Bin},
  journal={arXiv preprint arXiv:2410.08500},
  year={2024}
}

@article{liu2025toponav,
  title={TopoNav: Topological Graphs as a Key Enabler for Advanced Object Navigation},
  author={Liu, Peiran and Zhang, Qiang and Peng, Daojie and Zhang, Lingfeng and Qin, Yihao and Zhou, Hang and Ma, Jun and Xu, Renjing and Ji, Yiding},
  journal={arXiv preprint arXiv:2509.01364},
  year={2025}
}

@inproceedings{wang2023gridmm,
  title={Gridmm: Grid memory map for vision-and-language navigation},
  author={Wang, Zihan and Li, Xiangyang and Yang, Jiahao and Liu, Yeqi and Jiang, Shuqiang},
  booktitle={Proceedings of the IEEE/CVF International conference on computer vision},
  pages={15625--15636},
  year={2023}
}

@inproceedings{zhang2025novel,
  title={Mapnav: A novel memory representation via annotated semantic maps for vlm-based vision-and-language navigation},
  author={Zhang, Lingfeng and Hao, Xiaoshuai and Xu, Qinwen and Zhang, Qiang and Zhang, Xinyao and Wang, Pengwei and Zhang, Jing and Wang, Zhongyuan and Zhang, Shanghang and Xu, Renjing},
  booktitle={ACL},
  year={2025}
}

@inproceedings{li2024memonav,
  title={Memonav: Working memory model for visual navigation},
  author={Li, Hongxin and Wang, Zeyu and Yang, Xu and Yang, Yuran and Mei, Shuqi and Zhang, Zhaoxiang},
  booktitle={Proceedings of the IEEE/CVF Conference on Computer Vision and Pattern Recognition},
  pages={17913--17922},
  year={2024}
}

@inproceedings{liu2024volumetric,
  title={Volumetric environment representation for vision-language navigation},
  author={Liu, Rui and Wang, Wenguan and Yang, Yi},
  booktitle={Proceedings of the IEEE/CVF conference on computer vision and pattern recognition},
  pages={16317--16328},
  year={2024}
}

@inproceedings{DBLP:conf/acl/ChenLXCLW24,
  author       = {Jiaqi Chen and
                  Bingqian Lin and
                  Ran Xu and
                  Zhenhua Chai and
                  Xiaodan Liang and
                  Kwan{-}Yee Kenneth Wong},
  title        = {MapGPT: Map-Guided Prompting with Adaptive Path Planning for Vision-and-Language
                  Navigation},
  booktitle    = {ACL},
  year         = {2024},
}

@inproceedings{werby2024hierarchical,
  title={Hierarchical open-vocabulary 3d scene graphs for language-grounded robot navigation},
  author={Werby, Abdelrhman and Huang, Chenguang and B{\"u}chner, Martin and Valada, Abhinav and Burgard, Wolfram},
  booktitle={ICRA},
  year={2024}
}

@inproceedings{golabi2021intelligent,
  title={Intelligent and fuzzy UAV transportation applications in aviation 4.0},
  author={Golabi, Mahmoud and Nejad, Mazyar Ghadiri},
  booktitle={Intelligent and Fuzzy Techniques in Aviation 4.0: Theory and Applications},
  year={2021},
}

@article{zhao2024joint,
  title={Joint optimization of loading, mission abort and rescue site selection policies for UAV},
  author={Zhao, Xian and Wang, Xinlei and Dai, Ying and Qiu, Qingan},
  journal={Reliability Engineering \& System Safety},
  year={2024},
}

@article{asadzadeh2022uav,
  title={UAV-based remote sensing for the petroleum industry and environmental monitoring: State-of-the-art and perspectives},
  author={Asadzadeh, Saeid and de Oliveira, Wilson Jos{\'e} and de Souza Filho, Carlos Roberto},
  journal={Journal of Petroleum Science and Engineering},
  year={2022},
}

@inproceedings{zhu2025llava3d,
  title={Llava-3d: A simple yet effective pathway to empowering lmms with 3d capabilities},
  author={Zhu, Chenming and Wang, Tai and Zhang, Wenwei and Pang, Jiangmiao and Liu, Xihui},
  booktitle={ICCV},
  year={2025}
}

@inproceedings{jain2024odin,
  title={Odin: A single model for 2d and 3d segmentation},
  author={Jain, Ayush and Katara, Pushkal and Gkanatsios, Nikolaos and Harley, Adam W and Sarch, Gabriel and Aggarwal, Kriti and Chaudhary, Vishrav and Fragkiadaki, Katerina},
  booktitle={CVPR},
  year={2024}
}

@inproceedings{perez2018film,
  title={Film: Visual reasoning with a general conditioning layer},
  author={Perez, Ethan and Strub, Florian and De Vries, Harm and Dumoulin, Vincent and Courville, Aaron},
  booktitle={AAAI},
  year={2018}
}

@article{kim2025openvlaoft,
  title={Fine-tuning vision-language-action models: Optimizing speed and success},
  author={Kim, Moo Jin and Finn, Chelsea and Liang, Percy},
  journal={arXiv preprint arXiv:2502.19645},
  year={2025}
}

@inproceedings{li2023blip2,
  title={Blip-2: Bootstrapping language-image pre-training with frozen image encoders and large language models},
  author={Li, Junnan and Li, Dongxu and Savarese, Silvio and Hoi, Steven},
  booktitle={ICML},
  year={2023},
}

@article{kim2024openvla,
  title={Openvla: An open-source vision-language-action model},
  author={Kim, Moo Jin and Pertsch, Karl and Karamcheti, Siddharth and Xiao, Ted and Balakrishna, Ashwin and Nair, Suraj and Rafailov, Rafael and Foster, Ethan and Lam, Grace and Sanketi, Pannag and others},
  journal={arXiv preprint arXiv:2406.09246},
  year={2024}
}

@inproceedings{krantz2020beyond,
  title={Beyond the nav-graph: Vision-and-language navigation in continuous environments},
  author={Krantz, Jacob and Wijmans, Erik and Majumdar, Arjun and Batra, Dhruv and Lee, Stefan},
  booktitle={ECCV},
  year={2020},
}

@article{hart1968formal,
  title={A formal basis for the heuristic determination of minimum cost paths},
  author={Hart, Peter E and Nilsson, Nils J and Raphael, Bertram},
  journal={IEEE transactions on Systems Science and Cybernetics},
  year={1968},
}

@article{zhang2024navid,
  title={Navid: Video-based vlm plans the next step for vision-and-language navigation},
  author={Zhang, Jiazhao and Wang, Kunyu and Xu, Rongtao and Zhou, Gengze and Hong, Yicong and Fang, Xiaomeng and Wu, Qi and Zhang, Zhizheng and Wang, He},
  journal={arXiv preprint arXiv:2402.15852},
  year={2024}
}

@article{DBLP:journals/tmlr/dinov2,
  author       = {Maxime Oquab and
                  Timoth{\'{e}}e Darcet and
                  Th{\'{e}}o Moutakanni and
                  Huy V. Vo and
                  Marc Szafraniec and
                  Vasil Khalidov and others},
  title        = {DINOv2: Learning Robust Visual Features without Supervision},
  journal      = {Trans. Mach. Learn. Res.},
  year         = {2024},
}

@inproceedings{DBLP:siglip,
  author       = {Xiaohua Zhai and
                  Basil Mustafa and
                  Alexander Kolesnikov and
                  Lucas Beyer},
  title        = {Sigmoid Loss for Language Image Pre-Training},
  booktitle    = {ICCV},
  year         = {2023},
}

@article{touvron2023llama2openfoundation,
      title={Llama 2: Open Foundation and Fine-Tuned Chat Models}, 
      author={Hugo Touvron and Louis Martin and Kevin Stone and others},
      year={2023},
      journal={arXiv preprint arXiv:2307.09288},
}

@inproceedings{wang2025open,
  title={Open-Vocabulary Octree-Graph for 3D Scene Understanding},
  author={Wang, Zhigang and Su, Yifei and Li, Chenhui and Wang, Dong and Huang, Yan and Li, Xuelong and Zhao, Bin},
  booktitle={ICCV},
  pages={7037--7047},
  year={2025}
}

@inproceedings{radford2021learningclip,
  title={Learning transferable visual models from natural language supervision},
  author={Radford, Alec and Kim, Jong Wook and Hallacy, Chris and Ramesh, Aditya and Goh, Gabriel and Agarwal, Sandhini and Sastry, Girish and Askell, Amanda and Mishkin, Pamela and Clark, Jack and others},
  booktitle={International conference on machine learning},
  pages={8748--8763},
  year={2021},
  organization={PmLR}
}

@article{zhang2025citynavagent,
  title={CityNavAgent: Aerial Vision-and-Language Navigation with Hierarchical Semantic Planning and Global Memory},
  author={Zhang, Weichen and Gao, Chen and Yu, Shiquan and Peng, Ruiying and Zhao, Baining and Zhang, Qian and Cui, Jinqiang and Chen, Xinlei and Li, Yong},
  journal={arXiv preprint arXiv:2505.05622},
  year={2025}
}

@article{xu2025geonav,
  title={Geonav: Empowering mllms with explicit geospatial reasoning abilities for language-goal aerial navigation},
  author={Xu, Haotian and Hu, Yue and Gao, Chen and Zhu, Zhengqiu and Zhao, Yong and Li, Yong and Yin, Quanjun},
  journal={arXiv preprint arXiv:2504.09587},
  year={2025}
}
}

\end{document}